\documentclass{article} 
\usepackage{colm2024_conference}
\usepackage{microtype}
\usepackage{hyperref}
\usepackage{url}
\usepackage{booktabs}
\usepackage{dashrule}
\usepackage{arydshln}
\usepackage{graphicx}
\usepackage{pythonhighlight}
\definecolor{darkblue}{rgb}{0, 0, 0.5}
\hypersetup{colorlinks=true, citecolor=darkblue, linkcolor=darkblue, urlcolor=darkblue}

\usepackage{xcolor}

\colmfinalcopy
\title{Conceptual and Unbiased Reasoning in Language Models}

\author{Ben Zhou\textsuperscript{$\spadesuit$}\thanks{Most work done during Ben Zhou and Sihao Chen's internship at Tencent AI Lab.} \quad Hongming Zhang\textsuperscript{$\heartsuit$} \quad Sihao Chen\textsuperscript{$\spadesuit$} \\ \bf{Dian Yu\textsuperscript{$\heartsuit$} \quad Hongwei Wang\textsuperscript{$\heartsuit$} \quad Baolin Peng\textsuperscript{$\heartsuit$} \quad Dan Roth\textsuperscript{$\spadesuit$} \quad Dong Yu\textsuperscript{$\heartsuit$}} \\
\textsuperscript{$\spadesuit$}University of Pennsylvania\quad \textsuperscript{$\heartsuit$}Tencent AI Lab
}

\begin{document}

\maketitle

\begin{abstract}
Conceptual reasoning, the ability to reason in abstract and high-level perspectives, is key to generalization in human cognition. However, limited study has been done on large language models' capability to perform conceptual reasoning. In this work, we bridge this gap and propose a novel conceptualization framework that forces models to perform conceptual reasoning on abstract questions and generate solutions in a verifiable symbolic space. Using this framework as an analytical tool, we show that existing large language models fall short on conceptual reasoning, dropping 9\% to 28\% on various benchmarks compared to direct inference methods. We then discuss how models can improve since high-level abstract reasoning is key to unbiased and generalizable decision-making. We propose two techniques to add trustworthy induction signals by generating familiar questions with similar underlying reasoning paths and asking models to perform self-refinement. Experiments show that our proposed techniques improve models' conceptual reasoning performance by 8\% to 11\%, achieving a more robust reasoning system that relies less on inductive biases.
\end{abstract}

\section{Introduction}

Language models have progressed rapidly in recent years after community research into the emergent abilities in models with billions of parameters and trillions of pre-training tokens. While such models achieve admirable performances on many reasoning benchmarks, people have been looking into their decision-making process and trying to understand why large language models (LLM) perform well, especially compared with smaller models. One of the popular theories regarding why LLMs work is that they can memorize a huge amount of facts from pre-training data and generalize to new queries through induction on semantics \citep{tang2023large, Li2023Deceiving} (i.e., associate new inputs with memorized facts and guess the most likely outputs). Induction-based inference in LLMs works well for most daily tasks since these tasks are usually familiar to the model and contain strong induction signals. However, heavily relying on induction-based inference limits the model from performing well on non-induction tasks that require complex reasoning and planning, such as temporal reasoning \citep{zhou2020temporal} and travel planning \citep{xie2024travelplanner}. Moreover, relying on induction signals will inevitably lead to biased reasoning, where models perform much worse on queries containing unfamiliar semantics \citep{Li2023Deceiving}. 

A significant gap between LLM's induction-based reasoning and human reasoning, which does not have the issues above, is the ability to perform conceptual and abstract reasoning. Humans develop the ability to reason on a conceptual level from a very young age \citep{markovits2015metacognition, walker2015learning}, which enables them to overcome the limitations of induction through analogy and planning. For example, when humans travel to a new city, they can devise a plan to find good restaurants based on analogical experiences from the places they are familiar with because they can associate the underlying problem-solving processes on a conceptual level. However, there have been limited attempts to study LLMs' abstract reasoning because 1) it is difficult to force LLMs to perform abstract reasoning, and 2) it is challenging to verify the correctness of models' abstract reasoning process.

This work proposes a novel conceptualization framework to investigate and improve LLM's abstract reasoning capability. The conceptualization framework is composed of two parts, which address the challenges mentioned above respectively: 1) a question abstraction process that removes as much induction signal from the original question by replacing specific special nouns (e.g., named entities, numerical values) with their fine-grained semantic types; 2) a symbolic program space where LLMs will generate their abstract reasoning solutions in so that we can verify their correctness by executing the program with the necessary information from the original question and check for denotation accuracy. We use a state-of-the-art language model to perform the question abstraction step so that all downstream models work with a consistent set of high-quality abstract questions. Then, all downstream models are given the exact instructions regarding the design of the symbolic program space and try to generate solutions based on only the conceptualized question. In \S\ref{sec:analytical-results}, we show that existing LLMs' performances are much lower in the abstract reasoning setup, dropping 9\% to 28\% across different models compared to chain-of-thought (CoT) \citep{wei2022chain} that entirely rely on induction, indicating that existing LLMs lack the core reasoning capability to generalize on a conceptual level. This also suggests that while some models have seemingly close performances on various datasets, their underlying reasoning capability varies greatly as some gaps are dramatically larger than others.

We then propose improvement techniques for LLM conceptual reasoning. Our main motivation is finding high-quality induction signals and using them as a reference to the conceptual reasoning process. This approximates the human analogy processes that symbolically associate familiar problem-solving processes from past experiences. Specifically, we find concrete questions that can be reduced to the same conceptualized question but with familiar nouns. These questions are often directly answerable because the supporting evidence commonly appears in the pre-training data, and the LLM answer is more likely to be correct. We propose two ways to use such similar questions. They can be used as a run-time exam to identify high-performing candidate solutions, and we can infer final predictions based only on them. In addition, we can ask LLMs to self-refine the candidate programs based on CoT solutions to the mistakes they make. Experiments show the effectiveness of the proposed methods, which improve 8\% to 11\% on a wide range of reasoning benchmarks. Our proposed conceptual reasoning method achieves relatively comparable performances with CoT in some scenarios while being more robust and less biased because it produces symbolic solutions without heavily relying on induction signals and makes predictions based on deterministic executions of such symbolic solutions.

\section{Related Works}
Many works have identified that large language models
cannot handle tasks that require intensive reasoning~\citep{tang2023large, saparov2022language, feng2022generic, Li2023Deceiving, xu2023cognitive}. Our work builds on these works and identifies high-level and conceptual reasoning as a major bottleneck shared among these observations, limiting models from system-2 reasoning  \citep{kahneman2011thinking, weston2023system, lin2024swiftsage}. This work is also related to decomposition-based reasoning \citep{min2019multi, zhou2022learning}, and program-based inference methods \citep{chen2022program, gao2023pal, lu2024chameleon}, where these methods aim to shift from model performing probabilistic induction to symbolic deduction, similar to our motivations. This work is marginally relevant to other LLM self-revision works \citep{chen2023teaching, madaan2024self}, but we consider different signals.

\section{Conceptualization Framework}

\begin{figure}[h]
\begin{center}
    \includegraphics[scale=0.4]{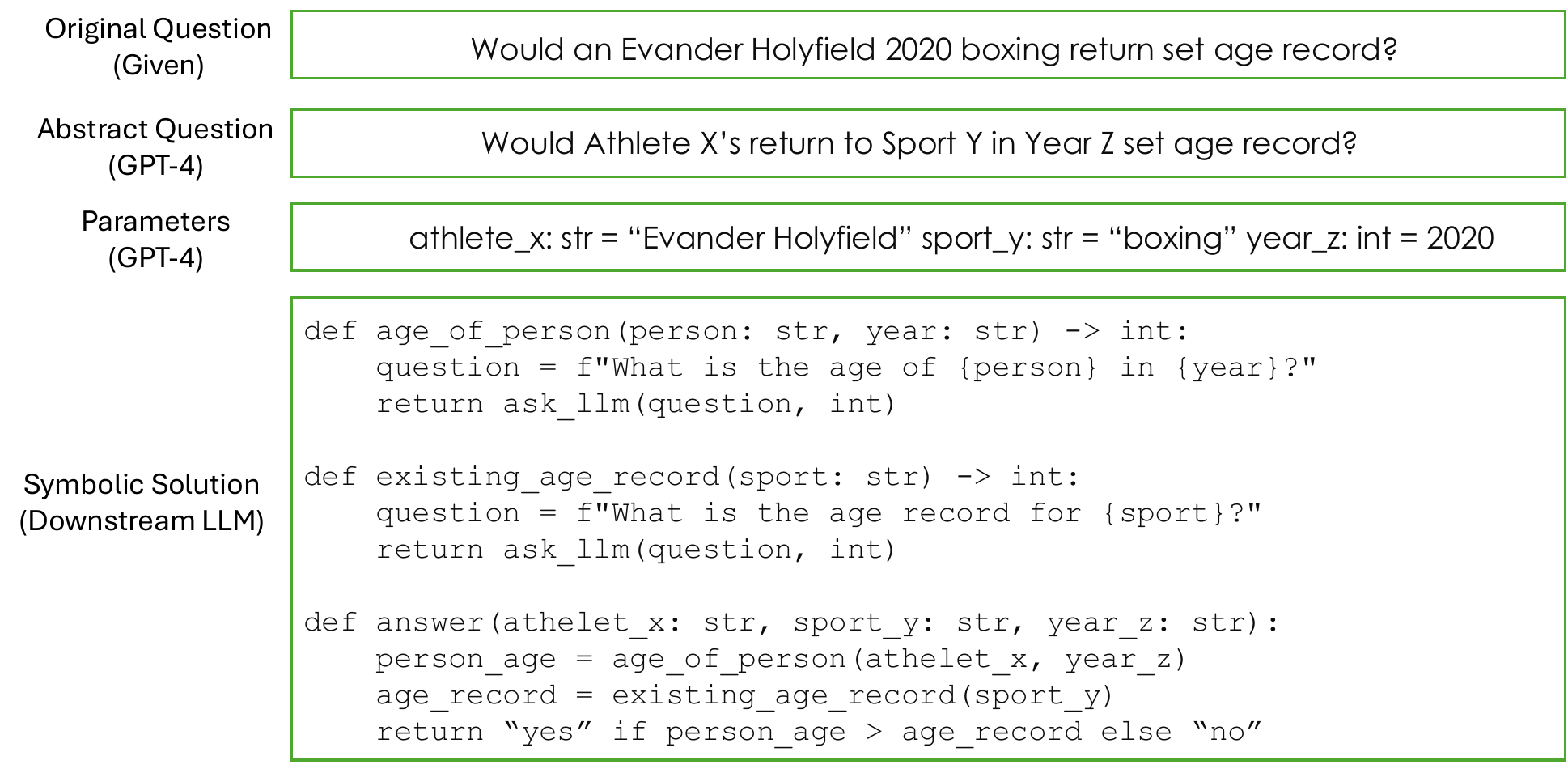}
    \caption{An overview of the conceptualization process. From an original question $Q$, GPT-4 generates an abstract question $Q_{abs}$ and corresponding parameters through a conceptualization process, and a downstream LLM will try to generate a symbolic solution based on only $Q_{abs}$, which is later executed with the actual parameters.}
    \label{fig:conceptualization-overview}
\end{center}
\end{figure}

This section introduces how the conceptualization framework is defined and implemented. For a natural question $Q$, we first employ a question conceptualization step that transforms $Q$ to $Q_{abs}$, an abstract version of the question. We then define a symbolic solution space using Python programs. At inference time, the downstream model generates a symbolic solution in our defined space based on only $Q_{abs}$, which is later deterministically executed with the concrete input parameters to generate a final prediction to the original question $Q$. This process is overviewed in Fig.~\ref{fig:conceptualization-overview}.

\subsection{Question Conceptualization}
\label{sec:conceptualization}
The goal of the question conceptualization step is to replace certain special nouns with their fine-grained semantic types so that humans and systems can develop a general solution that works for any specific nouns that fall under the semantic types. The primary challenge in this step is the tradeoff between solvability and generalizability. Even though all conceptualized questions are solvable with an unlimited program length (i.e., assuming a solution that considers all possible corner cases), it is often impractical and less solvable if we conceptualize too many concrete nouns in the question. On the other hand, if we remove too few nouns, models can still take the semantic shortcuts \citep{Li2023Deceiving} based on inductive biases. However, we emphasize that a balanced solution between these two tradeoffs is ideal but not required for our benchmark to be valid since models should, in theory, work on any level of conceptualization as long as the fine-grained semantic types are correct and the question is readable. We verify this later in \S\ref{sec:human-analysis}.

To balance this tradeoff, we use one of the most capable large language models, GPT-4, and prompt it with carefully designed few-shot examples that teach GPT-4 which nouns to replace and what fine-grained semantic types they should be replaced with. We use six demonstrations, selected manually from the StrategyQA \citep{geva2021did} dev set. We list the prompt in \S\ref{sec:prompts}. Given the original question $Q$, GPT-4 generates the abstract question $Q_{abs}$, a list of replaced noun phrases with their semantic types and corresponding Python types, and their concrete parameters representing the actual values from the original question $Q$. 

\subsection{Symbolic Program Space}
\label{sec:program-space}
We design a symbolic space based on Python programs to accommodate solutions for general binary (yes/no) questions. It includes several pre-defined helper functions to facilitate information retrieval and relational operations.

\subsubsection{Information Retrieval} 
We introduce a pre-defined helper function called $\mathrm{ask\_llm(query, type)}$ to handle all cases that require external knowledge. As the function name suggests, it accepts a string query that is often a short question, asks that question to the same downstream LLM that generates the program, and forces it to return the answer in the expected type. To ensure that the returned results are in the expected type, we employ a few-shot prompt that maps natural language queries to JSON structures, followed by a type casting. If the type casting fails, the retriever will retry ten times or raise an exception. We list this helper function with the few-shot examples in \S\ref{sec:prompts}. Because this design separates reasoning and knowledge, future works can adopt other retrieval methods with minimal effort. 

\subsubsection{Relational Operation Replacement} 
We implement a special post-processing step to handle mismatches in relational operators. For example, consider a program that needs to find the relationship between two people and check if they are parent-child. It will likely generate a query to the $\mathrm{ask\_llm(query, type)}$ helper function in the form of ``What is the relationship between person-1 and person-2?'', and the helper function may return multiple ways of describing the same relation, such as ``father-son'' or ``mother-daughter'', or ``descendent.'' Python built-in equal operator $==$ cannot handle such fuzzy matches, so the execution may not lead to the correct results.

\begin{table*}[ht]
\centering
\small
\begin{tabular}{lc}
\toprule
Operator & Query for Soft Operator \\
\cmidrule(lr){1-1}\cmidrule(lr){2-2}
a == b & Consider the implied value, is \textit{a} roughly the same as \textit{b}?\\
a $>$ b & Consider the implied value, is \textit{a} roughly larger than \textit{b}?\\
a $<$ b & Consider the implied value, is \textit{a} roughly smaller than \textit{b}?\\
a in b & Considered the implied value, is \textit{a} included or mentioned by the list \textit{b}? \\
\bottomrule
\end{tabular}
\caption{Queries to replace relational operators with soft operators. Other operations are compositions with the listed operations.}
\label{tab:soft-operator}
\end{table*}

To address this, we propose soft relational operators to replace built-in operators, including $>, >=, <, <=, ==, !=, in$, and $not\ in$. Each soft operator translates a compiler-level relational operation to a soft query to large language models. The specific queries that are used to replace the built-in operators are listed in Table~\ref{tab:soft-operator}. 
We implement actual helper functions that correspond to each operation. For example, we implement a helper function named $\mathrm{gt}(a, b)$ that calls $\mathrm{ask\_llm(\cdot)}$ with the queries for $>$ operations in Table~\ref{tab:soft-operator}. Then, we use Python's abstract syntax trees\footnote{\url{https://docs.python.org/3/library/ast.html}} and recursively find \textit{BoolOps} that fall under our interested relational operators and replace them with calls to our implemented helper functions. 

\subsection{Symbolic Program Execution}
As mentioned above, each question $Q$ is transformed to $Q_{abs}$ and a set of parameters with concrete values, as shown in Fig.~\ref{fig:conceptualization-overview}. In addition, each generated symbolic solution will have a function named $\mathrm{answer}(\cdot)$, which accepts the parameters as input values and returns a prediction string. At inference time, we prepend the actual implementations of our helper functions, including $\mathrm{ask\_llm(\cdot)}$ and all relational helper functions to the generated symbolic solution. We also append a function call to the $\mathrm{answer}(\cdot)$ function with the parameters from the original question and print the returned value from this function call. Finally, we execute this generated Python file and use the printed string as the final prediction of the model.

\subsection{Applications of the Conceptualization Framework}
We discuss two applications of the framework: evaluation and unbiased reasoning. 

\subsubsection{Analytical Evaluation}
\label{sec:analytical-evaluation}
Our proposed conceptualization framework can be an analytical tool to check if existing LLMs understand the problem-solving process or are simply taking semantic shortcuts \citep{Li2023Deceiving}. At inference time, a downstream model is provided with $Q_{abs}$, and few-shot demonstrations defining the available helper functions and program space. The model then generates K candidate programs, which are later executed with the corresponding input parameters and then used to majority vote a final answer based on the executed results. All downstream models are provided with the same $Q_{abs}$ and few-shot demonstrations and executed with the same input parameters.\footnote{$Q_{abs}$ and the input parameters are generated by GPT-4 as shown in Fig.~\ref{fig:conceptualization-overview}, and an expert writes the few-shot examples.} We detail the prompts used for generating candidate program solutions in \S\ref{sec:prompts}.

This inference process forces the model not to reason with memorized knowledge from particular nouns in the original question but to devise a solution based on the question's underlying reasoning path on an abstract level. Ideal reasoners should show relatively small gaps between performing conceptual reasoning and direct inferences (e.g., CoT).

\subsubsection{Unbiased Reasoning}
\label{sec:unbiased-reasoning}

\begin{figure}[h]
\begin{center}
    \includegraphics[scale=0.4]{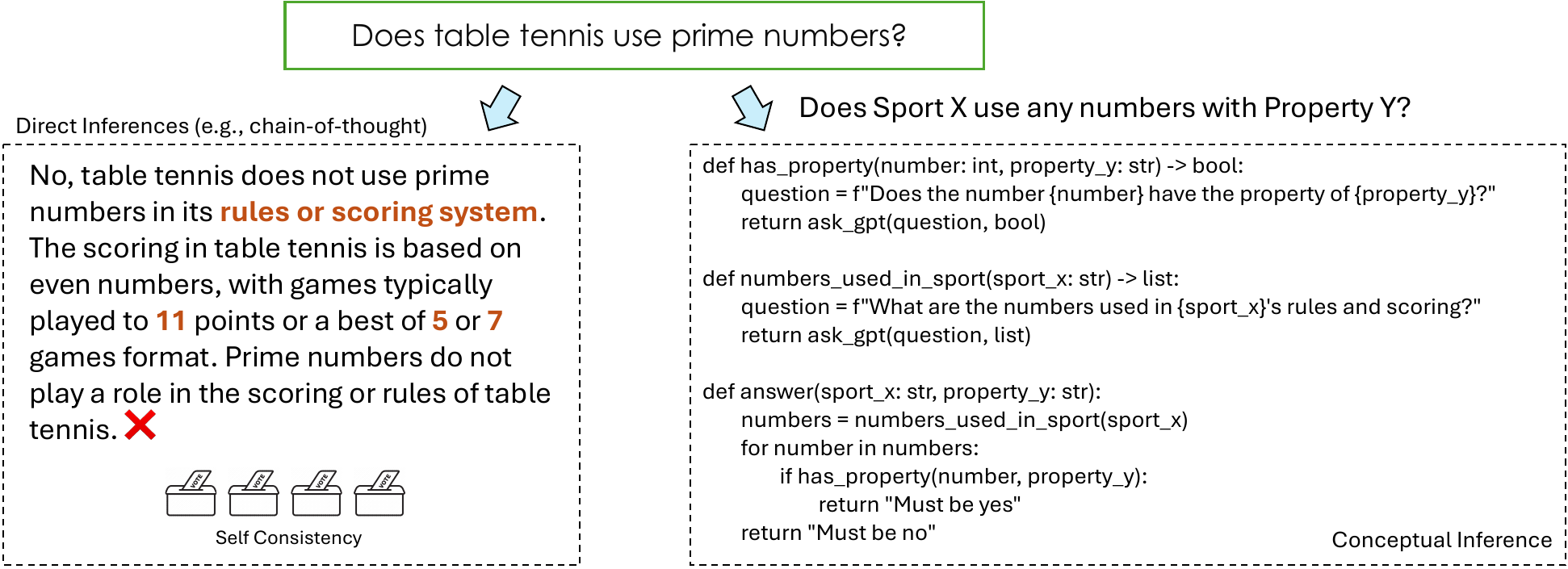}
    \caption{Comparison of a direct inference method (left) and conceptual inference (right). Results are taken from gpt-3-turbo.}
    \label{fig:reasoning-comparison}
\end{center}
\end{figure}

Taking one step further with the analytical settings, we can use the conceptualization framework to perform unbiased reasoning. This is based on observations that LLMs often make stubborn mistakes if certain word combinations strongly hint toward an incorrect answer. For example, as Fig.~\ref{fig:reasoning-comparison} shows, GPT-3.5 makes a stubborn mistake with direct inference on the left. This is because the model strongly memorizes prime numbers as a mathematical concept, so it cannot correctly associate that with sports. As a result, even though the model correctly identifies the need to find numbers in the sports' rules and scoring systems, it still cannot derive the correct answer. 

On the other hand, if the model follows our proposed conceptual inference process as shown on the right of Fig.~\ref{fig:reasoning-comparison}, it can correctly generate an unbiased solution since it cannot see the specific names of the sport and the numerical property, which can later be executed to find the correct answer. That being said, our proposed conceptualization framework can be applied for unbiased and neutral model reasoning, less affected by induction signals that might be irrelevant to problem-solving.

\section{Conceptual Reasoning with Analogy}
\label{sec:analogy}
As mentioned in \S\ref{sec:unbiased-reasoning}, the conceptualization framework can perform unbiased reasoning by removing most induction signals from the input question and forcing the model to perform abduction and deduction. This motivates us to discuss how we can improve the performance of conceptual reasoning and ideally reach the performances of direct inference methods, which have full access to induction signals such as entity names. 

This section proposes a method to inject \textbf{trustworthy} induction signals back to the reasoning process. Here, we emphasize that compared to a direct inference method that uses inductions from the original question, which are unreliable when the question differs much from pre-training distribution, our key motivation is to add induction signals familiar to the language model itself and, hence, more trustworthy. Specifically, for $Q_{abs}$, we generate a list of concrete questions with familiar nouns that can be conceptualized into the same abstract question. We use CoT to answer these questions and keep only those that can be confidently answered through self-consistency. 
This approximates the human analogy process: when we encounter a new situation, we often resort to familiar situations from personal experiences and use what we did back then to solve the current task.

In \S\ref{sec:similar-question-generation}, we introduce how we find these trustworthy similar questions that language models can accurately answer. In \S\ref{sec:program-selection}, we describe the first method of injecting such induction signals into the conceptual inference process, namely using them to select better candidate programs. In \S\ref{sec:self-refinement}, we introduce a self-refinement method using these questions as feedback.

\subsection{Similar Question Acquisition}
\label{sec:similar-question-generation}

\subsubsection{Question Generation}
To acquire the maximum diversity and correctness, we design a novel pipeline process to generate similar questions based on an abstract question $Q_{abs}$, with the goal of generating questions with similar underlying reasoning paths and solvable by LLMs directly. This similar question generation pipeline is performed with the same downstream LLM that generates the programs so that we ensure that the reasoning process does not involve a more capable model such as GPT-4. 

We first generate a list of concrete nouns/entities for each semantic type. For example, consider \textit{Is City X on Coast Y?}, the abstract version of \textit{Is Miami on the American West Coast?} we ask the downstream LLM to generate possible City X's. It will generate a list of city names familiar to the LLMs, such as Hong Kong, New York, and London. Similarly, it will generate a list of Coast Y's similar to \textit{American West Coast}, such as the Persian Gulf Coast, Baltic Coast, and the Black Sea Coast. During this process, we provide it with the nouns in the original question to ensure that the generated nouns are roughly in the same scope as the original entity. This way, the model will not generate a small coast such as Rockport Coast, which is too small to accommodate any cities. 

For each of the generated concrete nouns, we prompt the language model to generate a statement corresponding to either a positive answer or a negative answer to the original question.\footnote{This is because we only consider binary questions in this work.} For example, given the newly generated entity ``Hong Kong'' for City X, we design a few-shot prompt for the language model to generate positive statements with concrete Coast Y, such as \textit{Hong Kong is on China's southern coast}, as well as negative statements such as \textit{Hong Kong is not on the French Riviera}. We then transform these statements back to a question in the form of the original question, which is \textit{Is Hong Kong on China's southern coast?} and \textit{Is Hong Kong on the French Riviera?}, respectively. At the same time, the model will also generate corresponding input parameters, such as \textit{city\_x=``Hong Kong'', coast\_y=``China's southern coast''}. Prompts used in this step are listed in \S\ref{sec:prompts}.

\subsubsection{Question Validation}

\begin{table}[ht]
\centering
\small
\begin{tabular}{lcc}
\toprule
Agreement & Accuracy & \% Remaining \\
\cmidrule(lr){1-1}\cmidrule(lr){2-2}\cmidrule(lr){3-3}
None & 81.1 & 100 \\
8/10 & 89.7 & 73 \\
10/10 & 97.2 & 47 \\
\bottomrule
\end{tabular}
\caption{Sanity-check experiment on StrategyQA dev set. This compares the performance and remaining instances if we only evaluate instances where a certain number of predictions out of 10 samples agree. Experiments shown use Llama-70B-instruct and CoT.}
\label{tab:sanity-check-agreement}
\end{table}

We employ a verification process to select those that can be confidently answered and are likely to be more trustworthy by finding silver labels for each generated question with CoT. We run the chain-of-thought inference ten times for each question and keep the questions where at least 8 out of the 10 samples agree with each other.
To motivate this, we conduct a sanity-check experiment on StrategyQA with Llama-70B-instruct.\footnote{\url{https://huggingface.co/upstage/Llama-2-70b-instruct}} As Fig.~\ref{tab:sanity-check-agreement} shows, there is a significant improvement in performance if we only evaluate instances where CoT predictions agree with each other over a certain amount of times. This suggests that the labels we acquire after enforcing such an agreement are more reliable. On the other hand, an 8/10 agreement provides a good tradeoff between accuracy and the number of remaining instances, which will save computational costs. 

\subsection{Program Selection}
\label{sec:program-selection}
This section describes a straightforward method to apply these generated questions and their silver labels to improve model conceptual reasoning, namely a selection method that finds candidate programs that are likely to be better than the others. Specifically, within each question, we execute the K candidate programs generated in \S\ref{sec:analytical-evaluation} on all generated similar questions and evaluate each program's accuracy on the corresponding silver labels. We then rank the programs based on their accuracies and predict the original question based on the top programs' execution results. We employ a majority-vote process if multiple programs have the same accuracy. If there is a tie in the majority vote, we perform a weighted vote with all programs using their accuracies as the weight. 

\subsection{Program Self-refinement}
\label{sec:self-refinement}

\begin{figure}[h]
\begin{center}
    \includegraphics[scale=0.33]{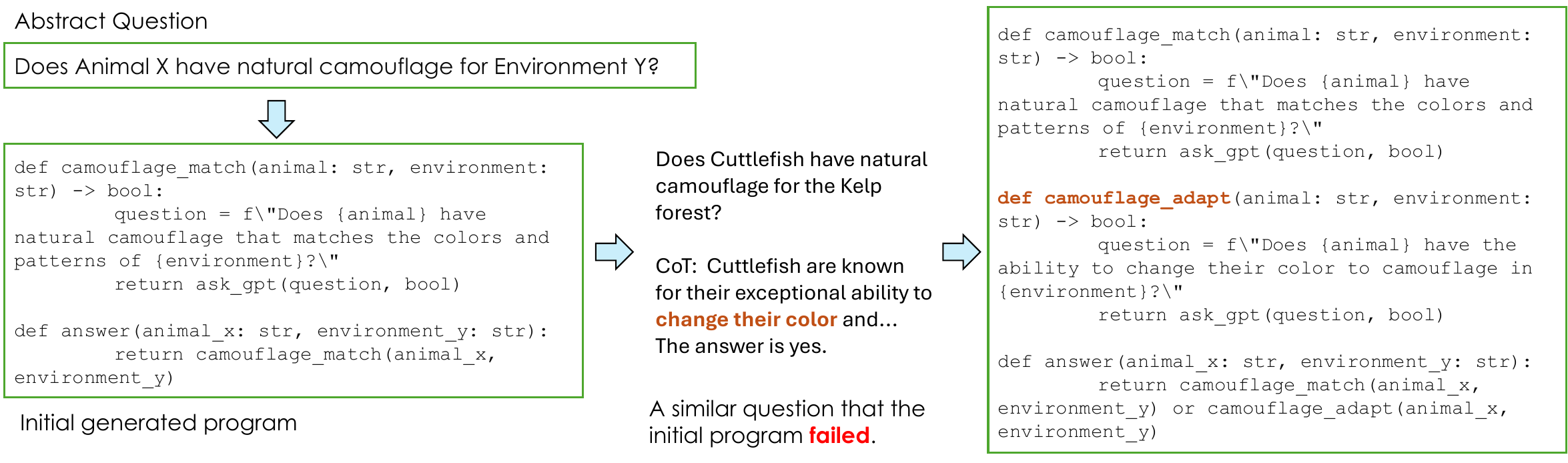}
    \caption{An overview of the self-refinement process with a concrete example. The initial program does not consider that some animals can change their color for camouflage purposes, as pointed out in the generate similar question's CoT solution. Based on this information, the model can refine the initial program to include such consideration.}
    \label{fig:self-refine}
\end{center}
\end{figure}

Using similar questions as a program selection method does not fully utilize their potential. This is because the reasoning processes from chain-of-thought are often correct on the generated similar questions because of the strong induction signal. As a result, if a program fails on certain similar questions, the CoT outputs serve as great feedback for LLMs to self-refine the candidate program. As overviewed in Fig.~\ref{fig:self-refine}, we propose a self-refinement process to ask LLMs to generate a better program based on an existing program and natural language feedback from CoT on why the existing program fails on a similar question. Because of computational limitations, we only conduct this process on programs with top accuracies within each question. We join the new programs with the original programs and conduct the same inference process described in \S\ref{sec:program-selection}. 

\section{Experiments}

\begin{table*}[ht]
\centering
\begin{tabular}{lccccccc}
\toprule
System & BoolQ & CsQA & Hotpot & Strat.dev & Strat.test & All & Exe.\%\\
\cmidrule(lr){1-1}\cmidrule(lr){2-2}\cmidrule(lr){3-3}\cmidrule(lr){4-4}\cmidrule(lr){5-5}\cmidrule(lr){6-6}\cmidrule(lr){7-7}\cmidrule(lr){8-8}
GPT-3.5 CoT & 72.6 & 60.3 & 85.3 & 75.0 & 63.3 & 71.9 & -\\
GPT-3.5 Prog. & 58.3 & 37.6 & 78.3 & 68.0 & 61.6 & 63.4 & 89 \\
\midrule
Llama CoT & 72.1 & 80.8 & 85.1 & 81.1 & 69.5 & 77.6 & -\\
Llama Prog. & 58.9 & 53.4 & 65.6 & 61.0 & 56.8 & 59.7 & 81 \\
\midrule
Mixtral CoT & 72.6 & 67.9 & 84.9 & 74.1 & 67.2 & 74.0 & - \\
Mixtral Prog. & 41.1 & 10.7 & 55.3 & 53.9 & 52.9 & 46.1 & 73\\
\bottomrule
\end{tabular}
\caption{Performance comparisons between CoT (uses original questions) and Prog(ram) (uses conceptualized questions and the proposed program space in \S\ref{sec:program-space}). Exe.\% is the program execution success rate. Unsuccessful executions are similar to CoT outputs that do not contain an answer, so they do not affect the fairness of comparison.}
\label{tab:perf-conceptualize}
\end{table*}

\begin{table*}[ht]
\centering
\begin{tabular}{lcccccc}
\toprule
System & BoolQ & CsQA & Hotpot & Strat.dev & Strat.test & Avg.$\Delta$  \\
\cmidrule(lr){1-1}\cmidrule(lr){2-2}\cmidrule(lr){3-3}\cmidrule(lr){4-4}\cmidrule(lr){5-5}\cmidrule(lr){6-6}\cmidrule(lr){7-7}
Llama Program & 58.9 & 53.4 & 65.6 & 61.0 & 56.8 & - \\ 
+ Selection & 65.0 & 76.1 & 72.1 & 68.4 & 60.0 & +9.2 \\
+ Self-refine & 68.0 & 78.6 & 73.9 & 71.1 & 61.0 & +11.4\\
\midrule
Mixtral Program & 41.1 & 10.7 & 55.3 & 53.9 & 52.9 & - \\ 
+ Selection & 53.3 & 30.8 & 60.1 & 56.1 & 52.1 & +7.7 \\
+ Self-refine & 55.3 & 38.0 & 64.0 & 58.8 & 54.8 & +11.4 \\
\bottomrule
\end{tabular}
\caption{System performances on reasoning datasets with proposed improvements. Selection refers to the selection method proposed in \S\ref{sec:program-selection}, and Self-refine refers to the self-refinement technique proposed in \S\ref{sec:self-refinement}. Program is the same as in Table~\ref{tab:perf-conceptualize}.}
\label{tab:perf-ours}
\end{table*}

In this section, we conduct experiments on both the analytical setting as described in \S\ref{sec:analytical-evaluation} and the reasoning setting as described in \S\ref{sec:unbiased-reasoning} with improvements described in \S\ref{sec:analogy}.

\subsection{Datasets, Models and Settings}
We consider four reasoning benchmarks that will benefit from symbolic solutions: BoolQ \citep{clark2019boolq}, CommonsenseQA \citep{talmor2018commonsenseqa}, HotpotQA \citep{yang2018hotpotqa}, and StrategyQA \citep{geva2021did}. We only use yes/no questions from HotpotQA. We convert multiple-choice questions in CommonsenseQA to binary questions by prompting Llama to transform the original question with its correct answer to a binary question with a gold answer of `yes', as detailed in \ref{sec:prompts}. We randomly sample 197, 234, and 456 questions from BoolQ, CommonsenseQA, and HotpotQA. StrategyQA has 228 questions in the dev set and 482 in the test set that can be successfully conceptualized.

For models, we consider three language models that have a relatively close performance on many popular benchmarks, namely GPT-3-Turbo~\citep{brown2020language}, Llama-70B-instruct~\citep{touvron2023llama} , and Mixtral-8x7B~\citep{jiang2024mixtral}.\footnote{We use gpt-3.5-turbo-0301 for GPT-3, \textit{upstage/Llama-2-70b-instruct} for llama, and \textit{mistralai/Mixtral-8x7B-Instruct-v0.1} for Mixtral. The last two models are taken from Huggingface and inferred with vLLM \citep{kwon2023efficient}.} 

All CoT inferences use the same set of few-shot prompts selected from StrategyQA. All experiments use a self-consistency K=10. We count draw cases (i.e., \#\textit{yes}=\#\textit{no}) as `unknown'. Temperature is 0.7, except for similar question generation, which uses 1.0 for diversity. All manual prompts and other parameters are tuned on samples from the StrategyQA dev set.

\subsection{Analytical Experiments}
\label{sec:analytical-results}
Table~\ref{tab:perf-conceptualize} shows our experiment results on three large language models' conceptual reasoning performance on four benchmarks. We see that all models drop performances using our conceptualization framework for analytical evaluation. Among these models, GPT-3.5 is the most consistent model, dropping 8.5\% on average, while Mixtral has a dramatic drop of 28\%, partially due to the lower execution success rate from poorly generated programs. These results show that although some language models claim to have on-par performances on many existing benchmarks, their core reasoning capabilities to not rely on induction are very different. This suggests that some models may show artificial high performances through imitation and memorization rather than actual generalization to reasoning tasks.

\subsection{Improvement Experiments}
Table~\ref{tab:perf-ours} shows the effectiveness of our proposed techniques in improving LLMs' conceptual reasoning capability. Our proposed analogical inference from similar questions improves 9.2\% with Llama and 7.7\% with Mixtral. Self-refinement further improves both models by over 11\%. Such improvements suggest that existing LLMs cannot efficiently make implicit analogies and find these trustworthy induction signals by themselves out of the box. We also observe that Llama achieves comparable performances on BoolQ and CommonsenseQA as CoT after self-refinement, indicating the potential to achieve high-performing, controllable, and unbiased reasoning. We include additional analysis in \S\ref{sec:more-analysis}.

\subsection{Human Analysis}
\label{sec:human-analysis}

\begin{table}[ht]
\centering
\small
\begin{tabular}{lcc}
\toprule
 & Correctness & Solvability \\
\cmidrule(lr){1-1}\cmidrule(lr){2-2}\cmidrule(lr){3-3}
StrategyQA / CommonsenseQA & 96 / 92 & 88 / 80 \\
\bottomrule
\end{tabular}
\caption{Human analysis on correctness (the percentage of questions that are correctly conceptualized by GPT-4), and solvability (the percentage of questions humans can solve with the first few interpretations of the abstract question).}
\label{tab:human}
\end{table}

We use two human analysis metrics to demonstrate that our conceptualized questions are reasonable for models to solve. Correctness is the percentage of questions where the semantic abstraction (i.e., the fine-grained semantic types) is correct. Solvability, on the other hand, measures the percentage of solvable questions with only the abstract form of the question. In \S\ref{sec:conceptualization}, we argued that as long as the correctness of semantic abstraction is guaranteed, models should, in theory, solve all questions, making solvability equal to correctness. 
However, in our analysis, a question is considered solvable if the human expert can generate a solution without considering too many corner cases.

One of the authors conducted such analysis with 50 questions sampled from StrategyQA and CommonsenseQA, representing two distinct domains: factual and commonsense. Table~\ref{tab:human} shows the results of our human analysis. Our conceptualization correctness is high, ranging from 92\% to 96\%. Solvability is slightly lower, but it only represents the percentage of questions that can be solved within the initial thoughts of our human expert. 
Both metrics are lower on CommonsenseQA because these questions contain fewer named entities.


\section{Conclusion}
We propose a novel conceptualization framework to investigate LLMs' ability to perform conceptual and high-level reasoning without the help of inductive biases from special nouns in the original question. Experiments show that existing LLMs perform much worse in conceptual reasoning than direct inference methods, dropping 9\% to 28\% across various models and benchmarks. To improve models' conceptual reasoning capability for unbiased inference, we propose two techniques that add trustworthy induction signals to the conceptual reasoning process, improving over 11\%. From an analytical perspective, our work suggests that models cannot sufficiently perform conceptual reasoning and efficiently make analogies to similar situations out of the box. To this end, our proposed improvement techniques that explicitly add back trustworthy analogical signals are shown to be effective, motivating future research on generalizable and unbiased reasoning and planning with minimum reliance on induction.

\bibliography{colm2024_conference}
\bibliographystyle{colm2024_conference}

\appendix

\section{Examples and More Analysis}
\label{sec:more-analysis}

\begin{table*}[ht]
\centering
\small
\begin{tabular}{p{0.08\linewidth} | p{0.9\linewidth}}
\toprule
Question & Would students at Marist have to petition to get a rowing team? \\
Abstract & Would students at School X have to petition to get a Sports Team Y? \\
Sim.Q 1 & Would students at Yale University have to petition to get a cheerleading squad? \\
Sim.Q 2 & Would students at NYU have to petition to get a Lacrosse Team? \\
Program 1 (20\% Acc) & \begin{python}
def students_preferences(school: str, sport: str) -> bool:
	question = f"Would students at {school} prefer to have a {sport} sports team?"
	preferences = ask_gpt(question, bool)
	return preferences
def school_budget(school: str) -> int:
	question = f"What is the budget of {school}?"
	budget = ask_gpt(question, int)
	return budget
def answer(school_x: str, sports_team_y: str):
	preferences = students_preferences(school_x, sports_team_y)
	if preferences:
		budget = school_budget(school_x)
		if budget >= 5000000:
			return "Must be yes"
	return "Must be no"
\end{python} \\
Program 2 (100\% Acc) & \begin{python}
def has_sports_team(school: str, sport: str) -> bool:
	question = f"Does {school} have a {sport} team?"
	team_exists = ask_gpt(question, bool)
	return team_exists
def popularity_of_sport(sport: str) -> str:
	question = f"How popular is the sport {sport}?"
	popularity = ask_gpt(question, str)
	return popularity
def answer(school_x: str, sports_team_y: str):
	if has_sports_team(school_x, sports_team_y):
		return "Must be no"
	school_popularity = popularity_of_sport(sports_team_y)
	if eq_override(school_popularity, 'very popular'):
		return "Must be yes"
	return "Must be no"
\end{python} \\
\bottomrule
\end{tabular}
\caption{An example showing how generated similar questions (with 7 other questions omitted) are helpful in identifying a better candidate program. The top program does not first check if the school already has the sports team, which would fail on the two similar questions as shown. The bottom program checks it achieves better accuracy on similar questions, and eventually gets the correct answer on the original question when executed with corresponding parameters.}
\label{tab:p8}
\end{table*}

\begin{table*}[ht]
\centering
\small
\begin{tabular}{p{0.08\linewidth} | p{0.9\linewidth}}
\toprule
Question & Did any country in Portuguese Colonial War take a general neutral role in WWII? \\
Abstract & Did any country in War X take a general neutral role in War Y? \\
Program & \begin{python}
def participating_countries_in_war(war: str) -> list:
	question = f"Which countries were involved in {war}?"
	countries = ask_gpt(question, list)
	return countries
def answer(war_x: str, war_y: str):
	war_x_countries = participating_countries_in_war(war_x)
	war_y_countries = participating_countries_in_war(war_y)
	for country_x in war_x_countries:
		if country_x in war_y_countries:
			return "Must be yes"
	return "Must be no"
\end{python} \\
Self-refined & \begin{python}
def participating_countries_in_war(war: str) -> list:
	question = f"Which countries were involved in {war}?"
	countries = ask_gpt(question, list)
	return countries
def neutral_country_in_war(country: str, war: str) -> bool:
	question = f"Was {country} neutral in {war}?"
	neutral = ask_gpt(question, bool)
	return neutral
def answer(war_x: str, war_y: str):
	war_x_countries = participating_countries_in_war(war_x)
	war_y_countries = participating_countries_in_war(war_y)
	for country_x in war_x_countries:
		if (in_override(country_x, war_y_countries) and neutral_country_in_war(country_x, war_y)):
			return "Must be yes"
	return "Must be no"
\end{python} \\
\bottomrule
\end{tabular}
\caption{An example generated program and self-refined program. The original candidate's solution is incorrect because it does not check the countries' neutrality. The self-refined program fixes this mistake based on generated similar questions and their silver labels. }
\label{tab:p9}
\end{table*}

In this section, we show a few examples of generated programs in our symbolic solution space. In Table~\ref{tab:p8}, we show that generated similar questions are helpful in selecting better programs, which leads to better overall performances in the ``+Selection'' setting of Table~\ref{tab:perf-ours}. In Table~\ref{tab:p9}, we show the effectiveness of self-refinement.

\section{Prompts}
\label{sec:prompts}

\begin{table*}[ht]
\centering
\small
\begin{tabular}{p{0.08\linewidth} | p{0.9\linewidth}}
\toprule
Prompt & Content \\
\cmidrule(lr){1-1}\cmidrule(lr){2-2}
system & Identify named entities, special nouns, or numerical values in the given question, then replace some of them with appropriate semantic types, so that the resulting abstract question is still answerable with the same general solution as the original question. Follow the the provided examples. \\
user & Does Rusev have to worry about human overpopulation in his homeland? \\
assistant & We first identify all named entities, special nouns and numerical values: Rusev, human overpopulation, his homeland. In order to keep the question answerable, we can only replace Rusev with Person X, because the original question's solution depends on knowing the issue in consideration is human overpopulation. Moreover, replacing "his homeland" will lead to too much ambiguity. As a result, we can replace "Rusev" to "Person X" (person$\_$x: str) So the question becomes Does Person X have to worry about human overpopulation in his homeland? With parameters person$\_$x="Rusev" \\
user & Can the Very Large Telescope observe the largest mountain on Earth? \\
assistant & We first identify all named entities, special nouns and numerical values: the Very Large Telescope, the largest mountain on Earth. We can replace these entities with fine-grained semantic types that keep the question answerable. The original question's general solution works as long as it knows that the Very Large Telescope is a Telescope on Earth, and the object in question is a geographical feature also on Earth.  As a result, we can replace "the Very Large Telescope" to "Telescope X" (telescope$\_$x: str) "the largest mountain on Earth" to "Geographical Feature Y" (geo$\_$feature$\_$y: str) So the question becomes Can Telescope X on Earth observe Geographical Feature Y on Earth? With parameters telescope$\_$x="Very Large Telescope", geo$\_$feature$\_$y="the largest mountain on Earth" \\
user & Will the Albany in Georgia reach a hundred thousand occupants before the one in New York? \\
assistant & We first identify all named entities, special nouns and numerical values: Albany in Georgia, a hundred thousand, the one in New York. The original question's solution can be generalized to checking if a city will reach a specific population before another city. This means that we can conceptualize the two cities in the original question with "City", and the number as Population. As a result, we can replace "Albany in Georgia" to "City X" (city$\_$x: str) "a hundred thousand occupants" to "Population Y" (population$\_$y: int) "the one in New York" to "City Z" (city$\_$z: str) So the question becomes Will City X reach Population Y before City Z? With parameters city$\_$x="Albany, Georgia", population$\_$y=100000, city$\_$z="Albany, New York" \\
user & Would the top of Mount Fuji stick out of the Sea of Japan? \\
assistant & We first identify all named entities, special nouns and numerical values: Mount Fuji, Sea of Japan. The generalized solution of the original question can be applied as long as we know Mount Fuji is a mountain (we cannot simply conceptulize it into geographical feature because the solution only works for features that point upwards such as a mountain), and Sea of Japan is a body of water (similarly, it needs to have depths.) As a result, we can replace "Mount Fuji" to "Mountain X" (mountain$\_$x: str) "the Sea of Japan" to "Body of Water Y" (body$\_$of$\_$water$\_$y: str) So the question becomes Would the top of Mountain X stick out of Body of Water Y? With parameters mountain$\_$x="Mount Fuji", body$\_$of$\_$water$\_$y="the Sea of Japan" \\
user & Did any country in Portuguese Colonial War take a general neutral role in WWII? \\
assistant & We first identify all named entities, special nouns and numerical values: Portuguese Colonial War, a general neutral role, WWII. The original solution checks for the involving countries in the first war, then checks if any of these countries took a general neutral role in the second war. We should not replace "a general nutral role" with its semantic type, because the solution would be vastly different for other roles. As a result, we can replace "Portuguese Colonial War" to "War X" (war$\_$x: str) "WWII" to "War Y" (war$\_$y: str) So the question becomes Did any country in War X take a general neutral role in War Y? With parameters war$\_$x="Portuguese Colonial War", war$\_$y="WWII" \\
user & Was Lil Jon's top ranked Billboard song a collaboration with a member of The Lox? \\
assistant & We first identify all named entities, special nouns and numerical values: Lil Jon, Billboard, The Lox. In order to keep the question answerable with the original problem-solving path, we should not replace 'Billboard', because the solution depends on knowing what kind of song the question is asking for. As a result, we can replace "Lil Jon" to "Artist X" (artist$\_$x: str) "The Lox" to "Group Y" (group$\_$y: str) So the question becomes Was Artist X's top ranked Billboard song a collaboration with a member of Group Y? With parameters artist$\_$x="Lil Jon", group$\_$y="The Lox" \\
\bottomrule
\end{tabular}
\caption{Prompts for question conceptualization.}
\label{tab:p1}
\end{table*}

\begin{table*}[ht]
\centering
\small
\begin{tabular}{p{0.08\linewidth} | p{0.9\linewidth}}
\toprule
Prompt & Content \\
\cmidrule(lr){1-1}\cmidrule(lr){2-2}
system & Solve the given question. Follow the way in the given examples to answer the questions. At the end, you must say if you have to guess an answer, the answer is yes/no. \\
user & Can the Very Large Telescope observe the largest mountain on Earth? \\
assistant & The Very Large Scope is in Chile. The largest mountain on Earth is mount Everest, which is in Nepal. The earth curvature is in between the two entities and there is not a line of sight. As a result, if I have to guess an answer, the answer is no. \\
user & Did Bill Nye vote for Franklin Delano Roosevelt? \\
assistant & Bill Nye was born in 1955. Franklin Delano Roosevelt's last election was in 1944. Bill Nye was not born nor alive at the time Roosevelt was running. As a result, if I have to guess an answer, the answer is no. \\
user & Does table tennis use prime numbers? \\
assistant & Table tennis plays to 11 points each game. 11 is a prime number. As a result, if I have to guess an answer, the answer is yes. \\
user & Would an Evander Holyfield 2020 boxing return set age record? \\
assistant & Evander Holyfield will turn 58 years old at the end of 2020. Steve Ward holds the world's oldest boxer title at age 59. 58 is younger than 59 and not enough to set the record. As a result, if I have to guess an answer, the answer is no. \\
user & Would the top of Mount Kilimanjaro stick out of the Sea of Japan? \\
assistant & The height of Mount Kilimanjaro is about 5,895 meters, while the Sea of Japan has a maximum depth of 3,742 meters. So, the top of Mount Kilimanjaro is taller than the maximum depth of the Sea of Japan. If we were to put Mount Kilimanjaro in the Sea of Japan, it will stick out. As a result, if I have to guess an answer, the answer is yes. \\
\bottomrule
\end{tabular}
\caption{Prompts for chain-of-thought inference.}
\label{tab:p2}
\end{table*}

\begin{table*}[ht]
\centering
\small
\begin{tabular}{p{0.08\linewidth} | p{0.9\linewidth}}
\toprule
Prompt & Content \\
\cmidrule(lr){1-1}\cmidrule(lr){2-2}
system & Answer the question in the expected type. Use your best educated guess or estimation if needed. Follow the provided examples and return a JSON dictionary with key 'answer'. \\
user & How many people today are related to Genghis Khan? (int) \\
assistant & {"answer": 35000000} \\
user & What is the profession of Michael Jackson? (str) \\
assistant & {"answer": singer} \\
user & Who has more than one Nobel Prize? (list) \\
assistant & {"answer": ["John Bardeen", "Frederick Sanger", "Linus Pauling", "Marie Curie"]} \\
user & Does anchors on KBS speak Arabic? (bool) \\
assistant & {"answer": false} \\
\bottomrule
\end{tabular}
\caption{Prompts for ask$\_$llm implementation.}
\label{tab:p3}
\end{table*}

\begin{table*}[ht]
\centering
\small
\begin{tabular}{p{0.08\linewidth} | p{0.9\linewidth}}
\toprule
Prompt & Content \\
\cmidrule(lr){1-1}\cmidrule(lr){2-2}
system & Follow the examples, for the given conceptualized question, generate some possible entities based on the type, and the provided example. Try to be diverse and creative. \\
user & Question: Was a person sold License X for Artwork Y ripped off? Entity: License X Example: Creative Commons License \\
assistant & MIT License General Public License BSD License CopyLeft License Artistic License Exclusive License VARA License END \\
user & Question: Is Technology X able to make Object Y? Entity: Object Y Example: adenovirus \\
assistant & prosthetics plastic bottles coffee smartphones contact lenses furniture computer software food END \\
\bottomrule
\end{tabular}
\caption{Prompts for entity generation.}
\label{tab:p4}
\end{table*}

\begin{table*}[ht]
\centering
\small
\begin{tabular}{p{0.08\linewidth} | p{0.9\linewidth}}
\toprule
Prompt & Content \\
\cmidrule(lr){1-1}\cmidrule(lr){2-2}
system & Follow the examples for the given conceptualized statement and generate the missing entities based on the type so that the given statement holds. Try to be diverse and creative. \\
user & Statement: If someone loves chocolate, they enjoy Compound Y. Entity: Company Y: str (e.g., capsaicin) \\
assistant & {"Compound Y": "Theobromine", "statement": "If someone loves chocolate, they enjoy Theobromine."} {"Compound Y": "Phenylethylamine", "statement": "If someone loves chocolate, they enjoy Phenylethylamine."} {"Compound Y": "Anandamide", "statement": "If someone loves chocolate, they enjoy Anandamide."} {"Compound Y": "Tryptophan", "statement": "If someone loves chocolate, they enjoy Tryptophan."} {"Compound Y": "Caffeine", "statement": "If someone loves chocolate, they enjoy Caffeine."} \\
user & Statement: Person X was present at the Year Y Met Gala. Entity: Person X: str (e.g., Bruce Lee), Year Y: int (e.g., 1964) \\
assistant & {"Person X": "Blake Lively", "Year Y": 2018, "statement": "Blake Lively was present at the 2018 Met Gala."} {"Person X": "Kendall Jenner", "Year Y": 2021, "statement": "Kendall Jenner was present at the 2021 Met Gala."} {"Person X": "Rihanna", "Year Y": 2017, "statement": "Rihanna was present at the 2017 Met Gala."} {"Person X": "Lady Gaga", "Year Y": 2019, "statement": "Lady Gaga was present at the 2019 Met Gala."} {"Person X": "Beyonce", "Year Y": 2015, "statement": "Beyonce was present at the 2015 Met Gala."} \\
\bottomrule
\end{tabular}
\caption{Prompts for positive/negative statement generation.}
\label{tab:p5}
\end{table*}

\begin{table*}[ht]
\centering
\small
\begin{tabular}{p{0.08\linewidth} | p{0.9\linewidth}}
\toprule
Prompt & Content \\
\cmidrule(lr){1-1}\cmidrule(lr){2-2}
system & Based on the given question, a person wrote a Python program. They used the following helper functions: ask\_gpt(query: str, type): ask a simple question to the knowledgeable GPT model and returns the answer in the desired type. However, the program this person wrote failed on some test cases. Given the program, and the failed test cases with corresponding reasons, write a new program that address any potential issues. \\
user & Original Question: Did Coach X (coach\_x: str) turn her chair around for Contestant Y (contestant\_y: str) on TV Show Z (tv\_show\_z: str)? Program the person wrote: ''' We need to consider if Contestant Y appeared on TV Show Z and if Coach X turned their chair around for Contestant Y. '''  def participant(contestant: str, tv\_show: str) -> bool: 	question = f"Did {contestant} participate in {tv\_show}?" 	participation = ask\_gpt(question, bool) 	return participation def chair\_turn(coach: str, contestant: str, tv\_show: str) -> bool: 	question = f"Did {coach} turn their chair around for {contestant} on {tv\_show}?" 	chair\_turned = ask\_gpt(question, bool) 	return chair\_turned def answer(coach\_x: str, contestant\_y: str, tv\_show\_z: str): 	participated = participant(contestant\_y, tv\_show\_z) 	if not participated: 		return "Must be no" 	chair\_turned = chair\_turn(coach\_x, contestant\_y, tv\_show\_z) 	return "Must be yes" if chair\_turned else "Must be no" Failed case 1: answer(coach\_x="Adam Levine", contestant\_y="John Legend", tv\_show\_z="The Voice") Failed reason 1: The concrete question in this case is 'Did Adam Levine turn her chair around for John Legend on The Voice?' Adam Levine and John Legend have both been coaches on "The Voice," but the premise of the question contains a misunderstanding of the show\'s format. "Turning chairs" is something coaches do during blind auditions to signal they want an artist to join their team. Since John Legend and Adam Levine are both coaches, they would not be auditioning and therefore would not have a chair turn situation between them.  As a result, if I have to guess an answer, the answer is no. However, the program returned results 'Yes'. \\
assistant & def check\_contestant(contestant: str, tv\_show: str) -> bool: 	question = f"Did {contestant} participate in {tv\_show} as a contestant?" 	participation = ask\_gpt(question, bool) 	return participation def check\_coach(judge: str, tv\_show: str) -> bool: 	question = f"Did {judge} participate in {tv\_show} as a coach?" 	participation = ask\_gpt(question, bool) 	return participation def chair\_turn(coach: str, contestant: str, tv\_show: str) -> bool: 	question = f"Did {coach} turn their chair around for {contestant} on {tv\_show}?" 	chair\_turned = ask\_gpt(question, bool) 	return chair\_turned def answer(coach\_x: str, contestant\_y: str, tv\_show\_z: str): 	is\_coach = check\_coach(coach\_x, tv\_show\_z) 	is\_contestant = check\_contestant(contestant\_y, tv\_show\_z) 	if not is\_coach or not is\_contestant: 		return "Must be no" 	chair\_turned = chair\_turn(coach\_x, contestant\_y, tv\_show\_z) 	return "Must be yes" if chair\_turned else "Must be no" \\
\bottomrule
\end{tabular}
\caption{Prompts for self-refinement. Another example is omitted.}
\label{tab:p6}
\end{table*}

\begin{table*}[ht]
\centering
\small
\begin{tabular}{p{0.08\linewidth} | p{0.9\linewidth}}
\toprule
Prompt & Content \\
\cmidrule(lr){1-1}\cmidrule(lr){2-2}
system & Based on the given question, write a Python program with some abstraction that solves the given question and all other similar questions that can be solved in a similar fashion. Think as comprehensively as possible, so that the program would work on any inputs. You can use the following helper function: ask\_llm(query: str, type) to ask a simple question to the knowledgeable GPT model and returns the answer in the desired type. \\
user & Is there a rapper whose real name is similar to Rapper X (rapper\_x: str)'s real name have more than N times (multiplier\_n: int) of Award Y (award\_y: str) than Rapper X (rapper\_x: str)? Multiple Choices: A) yes B) no
 \\
assistant  & \begin{python}
'''
We can first find a list of rappers with enough Award Y, then check if any of them share a similar name with Rapper X.
'''

def rapper_real_name(name: str) -> str:
	question = f"What is the real or legal name of rapper {name}?"
	real_name = ask_gpt(question, str)
	return real_name
def rapper_awards(name: str, award_name: str) -> int:
	question = f"How many {award_name} has the rapper {name} won?"
	num_awards = ask_gpt(question, int)
	return num_awards
def list_of_rappers_with_certain_awards(award_name: str, award_count: int) -> list:
	question = f"Give me a list of rappers who have won at least {str(award_count)} {award_name}."
	list_of_rappers = ask_gpt(question, list)
	return list_of_rappers
def name_similar(name_1: str, name_2: str) -> bool:
	question = f"Is the name '{name_1}' similar to the name '{name_2}'?"
	name_is_similar = ask_gpt(question, bool)
	return name_is_similar
def answer(rapper_x: str, award_y: str, multiplier_n: int):
	reference_person_awards_count = rapper_awards(rapper_x, award_y)
	target_award_count = int(multiplier_n * reference_person_awards_count)
	candidates = list_of_rappers_with_certain_awards(award_y, target_award_count)
	reference_person_real_name = rapper_real_name(rapper_x)
	for candidate_name in candidates:
		candidate_real_name = rapper_real_name(candidate_name)
		if name_similar(candidate_real_name, reference_person_real_name):
			return "Must be yes"
	return "Must be no"
#The program ends here.

\end{python} \\
\bottomrule
\end{tabular}
\caption{Prompts for program generation. Six other examples are omitted.}
\label{tab:p7}
\end{table*}

Table~\ref{tab:p1} lists the prompts used for question conceptualization, including the system message and few-shot examples. Table~\ref{tab:p2} lists the few-shot prompt used for chain-of-thought. All domains and tasks use the same set of examples because we only consider binary questions in this work. Table~\ref{tab:p3} lists the prompts to implement the information retrieval function ask\_llm($\cdot$), encouraging LLM to output answers in the expected type. Table~\ref{tab:p4} shows the prompt used for noun generation, which is the first step in the similar question generation process. Table~\ref{tab:p5} details the prompts to generate positive/negative statements, as described in \S\ref{sec:similar-question-generation}. Table~\ref{tab:p6} shows the prompts used during the self-refinement process, with another few-shot example omitted. Table~\ref{tab:p7} demonstrates how programs are generated, with only one few-shot example shown due to space limitations.

\end{document}